\RequirePackage{fix-cm}
\documentclass[twocolumn]{svjour3}          
\smartqed  
\usepackage{graphicx}
\usepackage{natbib}
\usepackage{multirow}
\usepackage{amsmath}
\usepackage{amssymb}
\usepackage{bbm}
\usepackage{bm}
\usepackage{booktabs}
\usepackage{array} 
\usepackage{blindtext}

\newcommand{\tableCellHeight}{1.1}
\newcommand{\tabstyle}[1]{
  \setlength{\tabcolsep}{#1}
  \renewcommand{\arraystretch}{\tableCellHeight}
  \centering
  \footnotesize
}

\usepackage{pifont}
\newcommand{\cmark}{\ding{51}}
\newcommand{\xmark}{\ding{55}}

\usepackage{xcolor}
\definecolor{citecolor}{HTML}{0071bc}
\usepackage[colorlinks,citecolor=citecolor]{hyperref}

\makeatletter
\renewcommand\paragraph{
  \@startsection{paragraph} 
  {4} 
  {\z@} 
  {.5em \@plus1ex \@minus.2ex} 
  {-.5em} 
  {\normalfont\normalsize\bfseries} 
}
\makeatother
\begin{document}
\sloppy

\title{Semi-Supervised Domain Generalization with \\ Stochastic StyleMatch
}


\author{Kaiyang Zhou         \and
        Chen Change Loy \and
        Ziwei Liu
}


\institute{Kaiyang Zhou \at
              S-Lab, Nanyang Technological University, Singapore \\
              \email{kaiyang.zhou@ntu.edu.sg}           
           \and
           Chen Change Loy \at
              S-Lab, Nanyang Technological University, Singapore \\
              \email{ccloy@ntu.edu.sg}
           \and
           Ziwei Liu \at
           S-Lab, Nanyang Technological University, Singapore \\
           \email{ziwei.liu@ntu.edu.sg}
}

\date{Received: date / Accepted: date}

\maketitle

\begin{abstract}
Ideally, visual learning algorithms should be generalizable, for dealing with any unseen domain shift when deployed in a new target environment; and data-efficient, for reducing development costs by using as little labels as possible. To this end, we study \emph{semi-supervised domain generalization (SSDG)}, which aims to learn a domain-generalizable model using multi-source, partially-labeled training data. We design two benchmarks that cover state-of-the-art methods developed in two related fields, i.e., domain generalization (DG) and semi-supervised learning (SSL). We find that the DG methods, which by design are unable to handle unlabeled data, perform poorly with limited labels in SSDG; the SSL methods, especially FixMatch, obtain much better results but are still far away from the basic vanilla model trained using full labels. We propose \emph{StyleMatch}, a simple approach that extends FixMatch with a couple of new ingredients tailored for SSDG: 1) stochastic modeling for reducing overfitting in scarce labels, and 2) multi-view consistency learning for enhancing domain generalization. Despite the concise designs, StyleMatch achieves significant improvements in SSDG. We hope our approach and the comprehensive benchmarks can pave the way for future research on generalizable and data-efficient learning systems. The source code is released at \url{https://github.com/KaiyangZhou/ssdg-benchmark}.
\end{abstract}

\section{Introduction}

\begin{figure*}[t]
    \centering
    \includegraphics[width=\textwidth]{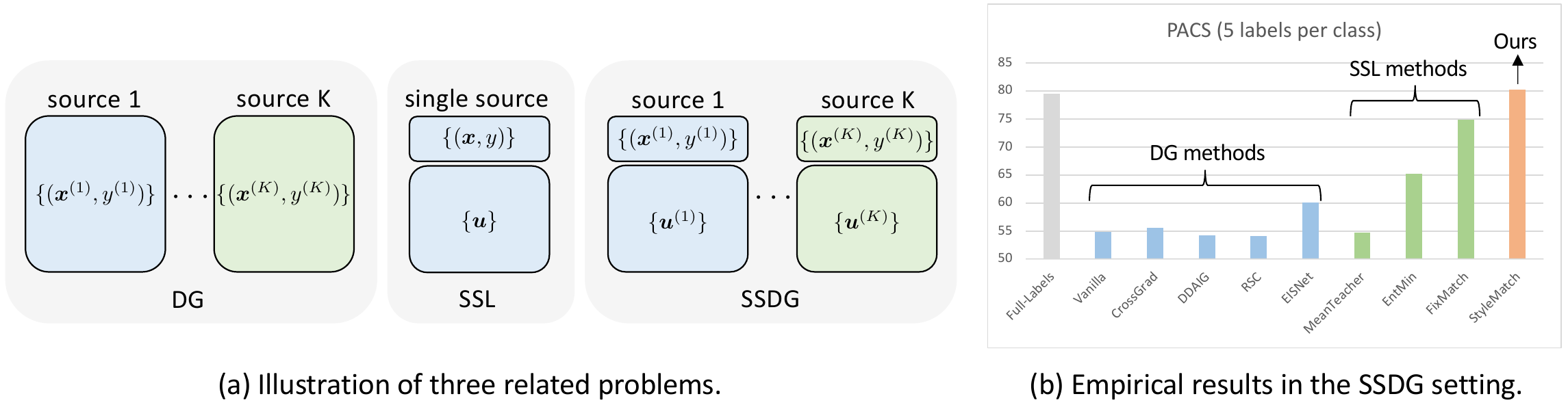}
    \caption{Semi-supervised domain generalization (SSDG) is closely related to domain generalization (DG) and semi-supervised learning (SSL), but poses unique challenges that cannot be solved by DG or SSL methods alone, i.e., multiple sources, domain shifts and partially-labeled training data.
    }
    \label{fig:intro_problems}
\end{figure*}

\begin{figure}
	\centering
	\includegraphics[width=\columnwidth]{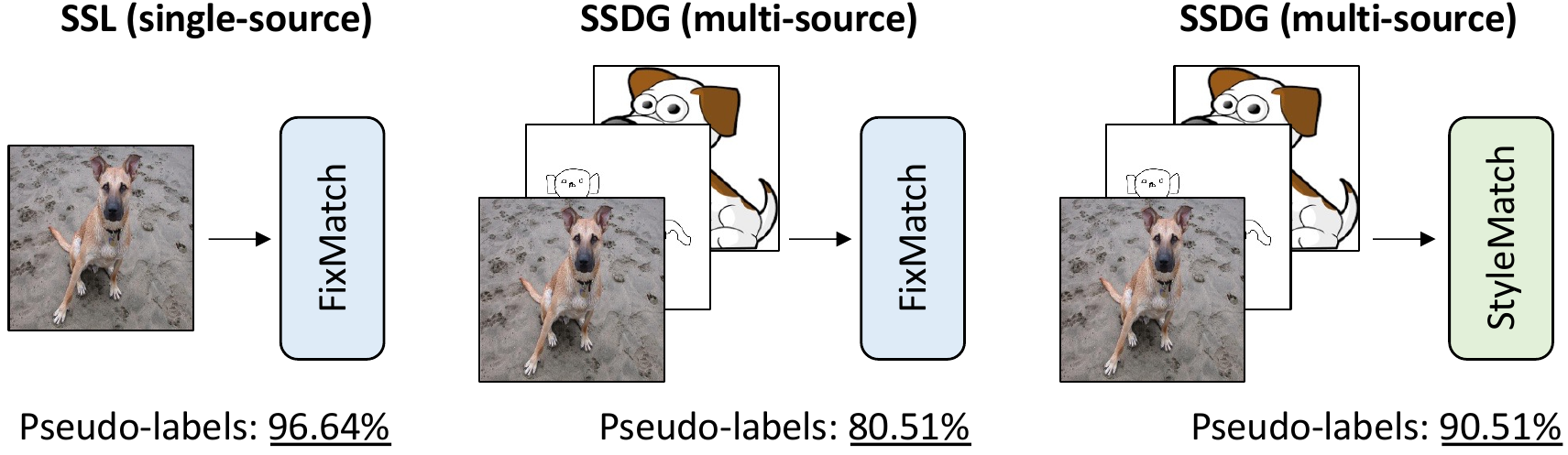}
	\caption{Directly applying FixMatch, which was developed for SSL, to SSDG is suboptimal as evidenced by the deterioration in pseudo-labeling accuracy. The proposed StyleMatch exhibits a better capability in handling unlabeled data in SSDG.}
	\label{fig:intro_pseudolabels}
\end{figure}

Visual recognition models are often deployed in a new target environment where the test data diverge significantly from the training data distribution---known as the domain shift problem~\citep{tzeng2015simultaneous,liu2020open,recht2019imagenet,zhou2021domain,li2017deeper,hoffman2018cycada}. For example, a dog classifier trained using only photo-realistic images might encounter images with a drastically different style at test time, such as sketch or cartoon, which the model has never seen during training and would thus incur huge performance drops. To overcome the domain shift problem, we would like to train a model in such a way that it becomes generalizable enough to handle any unseen domain shift. To this end, domain generalization (DG)~\citep{blanchard2011generalizing} is introduced, which studies how to develop domain-generalizable models using training data composed of a diverse set of sources, such as a combination of photo, sketch and cartoon images.

Besides the generalization capability, learning algorithms should ideally be data-efficient as well, meaning that we can train a model using as less labels as possible to lower down development costs. This topic is relevant to semi-supervised learning (SSL)~\citep{lee2013pseudo,grandvalet2004semi,tarvainen2017mean,sohn2020fixmatch,berthelot2019mixmatch}, which aims to exploit abundant unlabeled data along with limited labeled data for model training.

In this paper, we study \emph{semi-supervised domain generalization (SSDG)}, a new problem that considers both model generalization and data-efficiency under the same framework. Both DG and SSDG aim to learn models that can generalize to unseen target domains using only source data for training. However, DG has a strong assumption that all source domains are fully labeled. In contrast, SSDG adopts the SSL setting assuming only a handful of images within each source domain are labeled while a large quantity are unlabeled. A comparison of the three related problems is illustrated in Fig.~\ref{fig:intro_problems}(a).

To better understand the problem, we design two SSDG benchmarks based on two widely used DG datasets, and evaluate current state-of-the-art methods developed in the DG and SSL communities. A preview of the results is shown in Fig.~\ref{fig:intro_problems}(b). We find that the DG methods---which by design cannot use unlabeled data---obtain weak performance with limited labels. While the SSL methods, especially FixMatch~\citep{sohn2020fixmatch}, perform much better due to the use of unlabeled data, the gap with full-labels training---the pseudo oracle---is still noticeable. We also show in the experiments that a naive combination of DG and SSL methods does not fare well, either.

We address SSDG with a principled approach called \emph{StyleMatch}, which extends FixMatch with a couple of new ingredients tailored for SSDG. In particular, StyleMatch's designs are motivated to solve FixMatch's shortcomings identified by digging into the core component based on pseudo-labeling: as shown in Fig.~\ref{fig:intro_pseudolabels} (left vs middle), FixMatch works well in a single-source SSL scenario---which the algorithm is originally designed for---but suffers a substantial drop in accuracy when applied to a multi-source SSDG setting. We hypothesize that such a deterioration is caused by the shift in data distributions among different sources as well as the limited size of labels---challenges that are distinct to SSDG.

The first extension made in StyleMatch aims to prevent the model from overfitting small labeled data so that the model does not produce excessive wrong pseudo-labels with overconfidence. Specifically, we introduce uncertainty~\citep{gal2016bayesian,blundell2015weight} to the learning by modeling the classifier's weights with Gaussian distributions. This can be viewed as learning an ensemble of classifiers implicitly, which are sampled from the learnable distribution parameters.

The second extension is to convert FixMatch's two-view consistency learning framework into a multi-view version by adding style augmentation~\citep{huang2017arbitrary} as the third complementary view. Specifically, in addition to enforcing prediction consistency between weakly augmented images and the strongly augmented ones as in FixMatch, we further enforce prediction consistency between images from one domain and their style-transferred counterparts. Such a design has two benefits: first, the labeled group in each source domain is enlarged, which leads to more accurate pseudo-labels (see Fig.~\ref{fig:intro_pseudolabels}, right); second, the style augmentation-based learning essentially aligns the posterior probability among source domains, hence facilitating the learning of domain-invariant representations.

In summary, we make the following contributions:
\begin{enumerate}
	\item \emph{A new problem}: We study a new problem that takes into account model generalization and data-efficiency under the same framework, both critical for real-world vision applications but isolated in existing research.
	\item \emph{New benchmarks}: We design two benchmarks that cover a wide range of methods developed in relevant fields to facilitate future research.
	\item \emph{New insights}: We show that under the challenging SSDG setting, directly applying DG or SSL methods alone is insufficient, nor a naive combination of them.
	\item \emph{A new approach}: We propose a simple yet effective approach that seamlessly integrates two specifically designed components into an efficient pseudo-labeling method.
\end{enumerate}

\begin{figure*}[t]
    \centering
    \includegraphics[width=\textwidth]{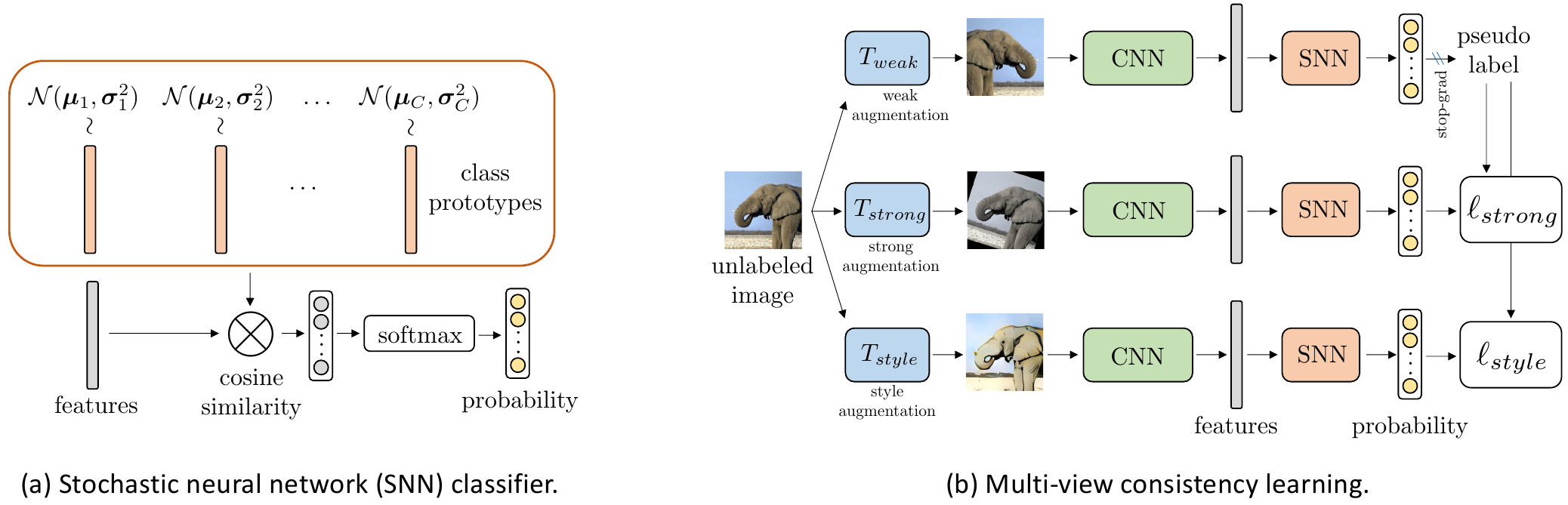}
    \caption{Two core components in StyleMatch. (a) The SNN-based design allows an ensemble of classifiers to be learned in an implicit manner. (b) The multi-view consistency learning framework is based on strong augmentation (geometric and color transformations) and style transfer (appearance and texture changes).
    }
    \label{fig:method}
\end{figure*}

\section{Related Work}

\subsection{Domain Generalization (DG)}
Many DG methods fall into the category of domain alignment, which aims to learn a domain-invariant feature space by aligning features across source domains. This is typically implemented by minimizing some distance measures such as first- and second-order moments~\citep{ghifary2017scatter}, maximum mean discrepancy (MMD)~\citep{li2018mmdaae}, and adversarial losses~\citep{li2018ciddg}. Meta-learning has also been extensively studied for DG~\citep{li2018learning,balaji2018metareg,feature_critic,dou2019domain,shu2021open}. Most meta-learning methods construct episodes by dividing source domains into a meta-train and a meta-test set without overlap, and learn a model on the meta-train set such that its performance on the meta-test set is improved.

Most related to our work are data augmentation methods, which can be generally categorized into four groups. The first group investigates traditional label-preserving transformations~\citep{volpi2019addressing}, such as adjusting image contrast and brightness. The second group is based on adversarial gradients for augmentation in the pixel space~\citep{shankar2018generalizing,volpi2018generalizing,qiao2020learning}. A representative work is CrossGrad~\citep{shankar2018generalizing}, which back-propagates adversarial gradients from a domain classifier to the input of a label classifier. The third family of methods also perform augmentation in the pixel space, but build the augmentation function using neural networks~\cite{zhou2020learning,xu2021robust,zhou2020deep}. For example, DDAIG developed by \citet{zhou2020deep} learns a neural network to transform images' appearance such that a domain classifier cannot identify their source domain labels. The last group transitions from pixel- to feature-level augmentation by, e.g., mixing feature statistics~\citep{zhou2021mixstyle} or learning feature perturbation networks~\citep{qiao2021uncertainty}.

Most existing DG methods cannot handle unlabeled data except those based on self-supervised learning~\citep{cvpr19jigen,wang2020learning}. The intuition behind using self-supervised losses for DG is to allow a model to discover patterns that are less correlated with class labels, hence reducing overfitting to source data. Nonetheless, our experiments show that the proposed approach fits the SSDG problem much better than previous DG methods. For a more comprehensive review in the DG area, we refer readers to the survey by \citet{zhou2021domain}.

\subsection{Semi-Supervised Learning (SSL)}
SSL is a well-established area with a plethora of methods developed in the literature. Most related to our work are those based on consistency learning~\citep{miyato2018virtual,tarvainen2017mean} and pseudo-labeling~\citep{sohn2020fixmatch,xie2020self}. The basic idea in consistency learning is to force a model's predictions on two different views of the same input to be similar to each other~\citep{zhou2004learning}. A consistency loss is often imposed on top-layer features~\citep{abuduweili2021adaptive} or the output probabilities~\citep{sohn2020fixmatch}. Recent studies have found that using a model's exponential moving average to generate the target for consistency learning can stabilize training~\citep{tarvainen2017mean}.

Pseudo-labeling~\citep{lee2013pseudo} provides either soft or hard pseudo-labels for unlabeled data using, e.g., a pretrained model~\citep{xie2020self} or the model being trained~\citep{sohn2020fixmatch}. Recent advances in pseudo-labeling~\citep{xie2020self,berthelot2019mixmatch,sohn2020fixmatch,xie2020unsupervised} have suggested that introducing strong noise to the student model can greatly improve performance, such as applying strong augmentation to the input or/and dropout to model parameters. To overcome distribution shifts between labeled and unlabeled data caused by sampling bias, a couple of studies~\citep{wang2019semi,abuduweili2021adaptive} have borrowed ideas from domain adaptation~\citep{hoffman2018cycada} to minimize feature distance.

SSDG is related to SSL as both need to deal with unlabeled data. However, in SSDG the unlabeled data are much more challenging to handle than those in SSL since \emph{the former are collected from heterogeneous sources}. Our experiments show that simply applying SSL methods to the SSDG problem is suboptimal.

\subsection{Stochastic Neural Networks}
Our work is also related to research in stochastic neural networks (SNNs), also known as Bayesian deep learning~\citep{gal2016bayesian}. The key idea is to drive exploration in the parameter space via stochastic modeling, i.e., casting weights as probability distributions~\citep{blundell2015weight}.

\citet{blundell2015weight} show that modeling neural networks' weights with Gaussian distributions allows the model to produce more reasonable predictions on a regression task with noisy data. \citet{gal2016bayesian} use Bernoulli distributions to model convolutional kernels, which are efficiently implemented by applying dropout at both training and test time~\citep{srivastava2014dropout}. From an application perspective, SNNs have been successfully applied to domain adaptation~\citep{lu2020stochastic}, semantic segmentation~\citep{kendall2017bayesian}, and person re-identification~\citep{yu2019robust}. However, \emph{their application in the DG area has not been identified before, and we are the first to successfully apply SNNs to SSDG}.

\section{Problem Definition}
In this section, we formally define the semi-supervised domain generalization (SSDG) problem. Let $\mathcal{X}$ and $\mathcal{Y}$ denote the input and label space respectively, a domain is defined as a joint distribution $P(X, Y)$ over $\mathcal{X} \times \mathcal{Y}$. $P(X)$ and $P(Y)$ denote the marginal distribution of $X$ and $Y$, respectively. In this work, we only consider distribution shifts in $P(X)$ while $P(Y)$ remains the same, i.e., all domains share the same label space.

Similar to the conventional DG setting, we have access to $K$ distinct but related source domains $\mathcal{S} = \{S_k\}_{k=1}^K$, each associated with a joint distribution $P^{(k)}(X, Y)$. Note that $P^{(k)}(X, Y) \neq P^{(k')}(X, Y)$ for $k \neq k'$. In SSDG, only a small portion of the source data have labels while many of them are unlabeled for which we can only access the (empirical) marginal distribution $P(X)$. For each source domain $S_k$, the labeled part is defined as $S_k^L = \{(\bm{x}^{(k)}, y^{(k)})\}$ with $(\bm{x}^{(k)}, y^{(k)}) \sim P^{(k)}(X, Y)$, and the unlabeled part as $S_k^U = \{\bm{u}^{(k)}\}$ with $\bm{u}^{(k)} \sim P^{(k)}(X)$. The size of unlabeled data is much bigger than that of labeled data, i.e., $|S_k^U| \gg |S_k^L|$.

The goal in SSDG is to learn a domain-generalizable model using both labeled and unlabeled source data. At test time, the model is directly deployed in an \emph{unseen} target domain $\mathcal{T} = \{\bm{x}^*\}$ with $\bm{x}^* \sim P^*(X)$. The target domain differs from any source domain, $P^*(X, Y) \neq P^{(k)}(X, Y)$ for $k \in \{1, ..., K\}$.

\section{Methodology}
Our StyleMatch is a simple approach based on assigning pseudo-labels to unlabeled data and using uncertainty and data augmentation to tackle issues caused by domain shifts. The architecture consists of two parts: 1) a neural network-based feature extractor, which takes images as input and produces vectorized feature representations, and 2) a linear classifier. In this work, we construct the feature extractor using a CNN.

The pseudo-labeling part is based on FixMatch~\citep{sohn2020fixmatch}, a state-of-the-art SSL method that forces predictions made on strongly augmented images to match pseudo-labels estimated using weakly augmented images.

The other two core components, i.e., stochastic classifier and multi-view consistency learning, are illustrated in Fig.~\ref{fig:method}. Below we detail the designs of these two components.

\subsection{Stochastic Classifier}
The main idea of designing a stochastic classifier is to optimize an ensemble of classifiers \emph{implicitly} during training for reducing overfitting~\citep{zhou2012ensemble}. We start by discussing a standard linear classifier. Let $\bm{z} \in \mathbb{R}^D$ denote a $D$-dimensional feature vector of image $\bm{x}$ and $C$ the total number of classes, a linear classifier with weights $\bm{W} \in \mathbb{R}^{C \times D}$ and biases $\bm{b} \in \mathbb{R}^C$ can be formulated as $\bm{W} \bm{z} + \bm{b}$. By ignoring the bias vector---which is common in recent contrastive learning methods~\citep{he2020momentum,chen2020simple}---we can view the weight matrix $\bm{W} = [\bm{w}_1, ..., \bm{w}_C]^T$, where $\bm{w}_c \in \mathbb{R}^D$ and $c = 1, ..., C$, as containing a set of class prototypes~\citep{snell2017prototypical}. From this perspective, the matrix-vector multiplication $\bm{W} \bm{z}$ essentially computes the (cross-correlation) similarity between the image $\bm{x}$ and each class prototype $\bm{w}_c$.

In our stochastic classifier, each class prototype is modeled using a Gaussian distribution parameterized by $\mathcal{N}(\bm{\mu}_c, \bm{\sigma}^2_c)$, where $\bm{\mu}_c, \bm{\sigma}_c \in \mathbb{R}^D$. At each training step, we sample for each class the prototype vector from the corresponding probability distribution, $\bm{w}_c \sim \mathcal{N}(\bm{\mu}_c, \bm{\sigma}^2_c)$. To allow end-to-end optimization, we employ a reparameterization trick~\citep{kingma2014auto,blundell2015weight} to bypass the discrete sampling process,
\begin{equation} \label{eq:reparam_trick}
\bm{w}_c = \bm{\mu}_c + \operatorname{softplus}(\bm{\sigma}_c) \odot \bm{\epsilon} \quad \text{where} \quad \bm{\epsilon} \sim \mathcal{N}(\bm{0}, \bm{I}).
\end{equation}

Once all class prototypes are obtained, the similarity scores are computed based on cosine similarity (denoted by $\operatorname{sim}(\cdot, \cdot)$), which are then passed to the softmax function for generating a normalized probability distribution,
\begin{equation} \label{eq:prob}
p(y|\bm{x}) = \frac{\exp(\operatorname{sim}(\bm{z}, \bm{w}_y)/\tau)}{\sum_{c=1}^C \exp(\operatorname{sim}(\bm{z}, \bm{w}_{c})/\tau)},
\end{equation}
where $\tau$ is a temperature hyper-parameter fixed to 0.05. See Fig.~\ref{fig:method}(a) for a sketch of the classifier.

Since the uncertainty parameters (i.e., the standard deviations) converge to small values, meaning that the sampled classifiers become more and more similar to each other toward the end of training~\citep{blundell2015weight,lu2020stochastic,yu2019robust}, we simply use the mean parameters (i.e., $\bm{w}_{c} = \bm{\mu}_c$) to classify images at test time.

\begin{figure*}[t]
    \centering
    \includegraphics[width=.85\textwidth]{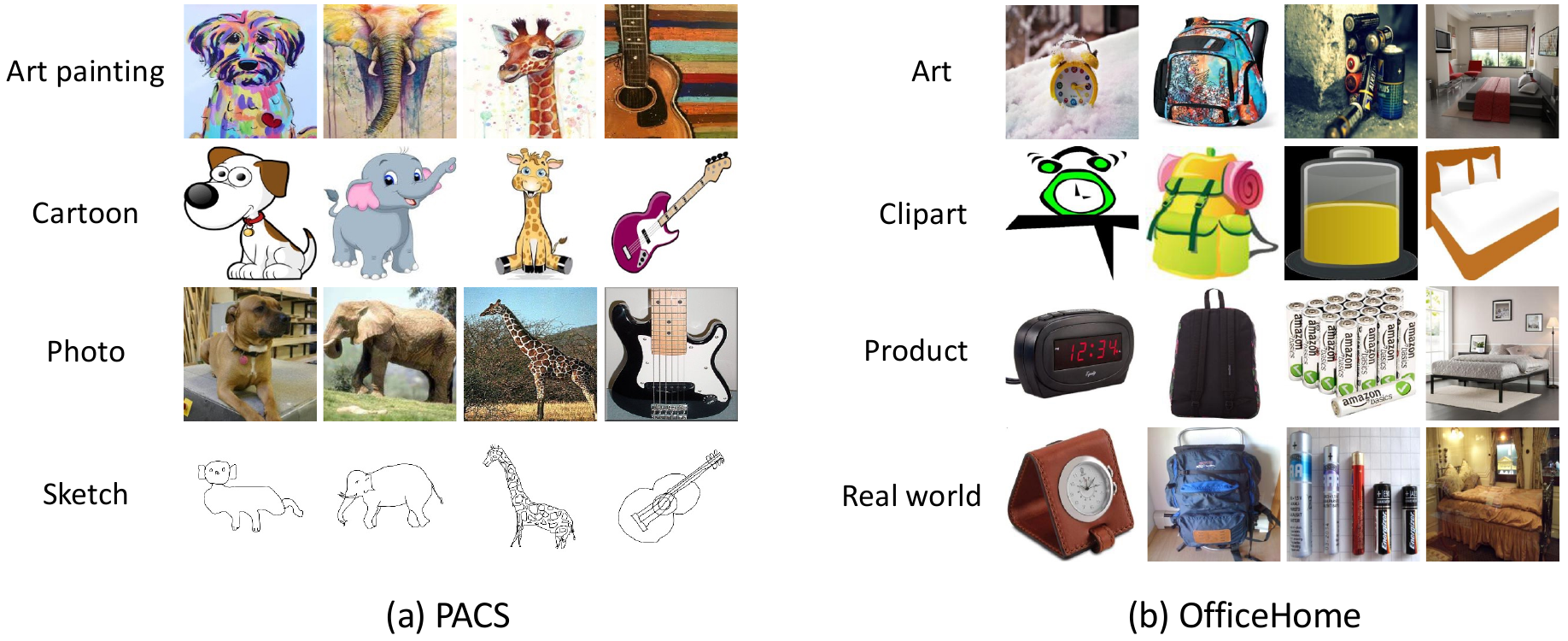}
    \caption{Example images from PACS~\citep{li2017deeper} and OfficeHome~\citep{office_home}, each consisting of four domains with drastically different image statistics. PACS mainly concerns image style changes while OfficeHome contains more sophisticated domain shifts like changes in image style, viewpoint, background, etc.}
    \label{fig:dataset_images}
\end{figure*}

\subsection{Multi-view Consistency Learning}
Here we discuss how to train the model using a multi-source, partially-labeled dataset. We propose an efficient learning framework called multi-view consistency learning to deal with unlabeled images and domain shifts. The framework is an extension of FixMatch~\citep{sohn2020fixmatch} and sketched in Fig.~\ref{fig:method}(b). There are three losses: one labeled loss and two unlabeled losses.

\paragraph{Labeled Loss}
For labeled source images, we compute the cross-entropy loss, defined as
\begin{equation} \label{eq:loss_labeled}
\ell_{labeled} = - \log p( y | T_{weak}(\bm{x}) ),
\end{equation}
where $(\bm{x}, y)$ is an image-label pair; $T_{weak}(\cdot)$ denotes a weak augmentation function, such as random crop and flip; $p( y | T_{weak}(\bm{x}) )$ is computed using Eq.~\eqref{eq:prob}.

\emph{For simplicity, the domain index is omitted and the loss formulation is based on an individual image (same below)}.

\paragraph{Unlabeled Losses}
Following FixMatch, given an unlabeled image $\bm{u}$, we first forward the weakly augmented version $T_{weak}(\bm{u})$ to the model to produce a probability distribution $q(\bm{u}) \in \mathbb{R}^C$ and a pseudo-label $\hat{q}(\bm{u}) = \arg \max q(\bm{u})$. Then, we force the model's prediction on the strongly augmented version $T_{strong}(\bm{u})$ to match the pseudo-label,
\begin{equation} \label{eq:loss_strong}
\ell_{strong} = - \mathbbm{1}(\max( q(\bm{u}) ) \geq \pi) \log p( \hat{q}(\bm{u}) | T_{strong}(\bm{u}) ),
\end{equation}
where $\mathbbm{1}(\cdot)$ is an indicator function and $\pi$ is a confidence threshold. We follow \citet{sohn2020fixmatch} to set $\pi = 0.95$ and compute this loss (and the other unlabeled loss) for both labeled and unlabeled images.

The strong augmentation function $T_{strong}(\cdot)$ is based on RandAugment~\citep{cubuk2020randaugment} and Cutout~\citep{devries2017cutout}.

To complement strong augmentation that is mainly related to geometric and color intensity-based transformations, we add a third view based on style transfer~\citep{huang2017arbitrary}. Specifically, we enforce prediction consistency for the same image but of different styles (domains). The motivation is two-fold: 1) to augment the labeled data within each source so the model can be better learned to produce more accurate pseudo-labels for the unlabeled data; 2) such a design aligns the posterior probability among source domains, i.e., $p(y|\bm{x}^{(k)}) = p(y|\bm{x}^{(k')})$ where $k \neq k'$, hence facilitating the learning of \emph{domain-invariant representations}. The loss is defined as
\begin{equation} \label{eq:loss_style}
\ell_{style} = - \mathbbm{1}(\max( q(\bm{u}) ) \geq \pi) \log p( \hat{q}(\bm{u}) | T_{style}(\bm{u}) ),
\end{equation}
where $T_{style}$ is implemented using AdaIN~\citep{huang2017arbitrary}. Note that $T_{style}(\bm{u})$ maps $\bm{u}$ to a different source domain than itself.

\paragraph{Final Loss}
The final learning objective is a combination of Eq.~\eqref{eq:loss_labeled}, \eqref{eq:loss_strong} and \eqref{eq:loss_style},
\begin{equation}
\ell_{all} = \ell_{labeled} + \ell_{strong} + \ell_{style}.
\end{equation}
We do not need any balancing weights for these losses.

\begin{table*}[t]
    \tabstyle{4pt}
    \caption{
    Domain generalization results in the low-data regime on PACS. A: Art painting. C: Cartoon. P: Photo. S: Sketch. $\bm{u}$: use unlabeled source data.
    }
    \label{tab:main_results_pacs}
    \begin{tabular}{l | c | cccc >{\em}c | cccc >{\em}c}
    \toprule
    \multirow{2}{*}{Model} & \multirow{2}{*}{$\bm{u}$} & \multicolumn{5}{c|}{\# labels: 210 (\emph{10 per class})} & \multicolumn{5}{c}{\# labels: 105 (\emph{5 per class})} \\
    & & A & C & P & S & {Avg} & A & C & P & S & {Avg} \\
    \midrule
    Full-Labels & - & 76.95 & 75.90 & 95.96 & 69.20 & {79.50} & 76.95 & 75.90 & 95.96 & 69.20 & {79.50} \\
    \midrule
    \multicolumn{12}{c}{\emph{Domain generalization methods}} \\
    Vanilla & \xmark & 63.09 & 58.49 & 86.56 & 45.56 & {63.42} & 56.71 & 53.87 & 71.87 & 36.92 & {54.84} \\
    CrossGrad & \xmark & 62.56 & 58.92 & 85.81 & 44.11 & {62.85} & 56.39 & 55.11 & 72.61 & 38.08 & {55.55} \\
    DDAIG & \xmark & 61.95 & 58.74 & 84.44 & 47.48 & {63.15} & 55.09 & 52.31 & 70.53 & 38.89 & {54.20} \\
    RSC & \xmark & 65.13 & 56.65 & 86.18 & 47.90 & {63.96} & 55.32 & 48.08 & 72.15 & 40.72 & {54.07} \\
    EISNet & \cmark & 66.84 & 61.33 & 89.16 & 51.38 & {67.18} & 62.08 & 54.75 & 80.66 & 42.68 & {60.04} \\
    \midrule
    \multicolumn{12}{c}{\emph{Semi-supervised learning methods}} \\
    MeanTeacher & \cmark & 62.41 & 57.94 & 85.95 & 47.66 & {63.49} & 56.00 & 52.64 & 73.54 & 36.97 & {54.79} \\
    EntMin & \cmark & 72.77 & 70.55 & 89.39 & 54.38 & {71.77} & 67.01 & 65.67 & 79.99 & 47.96 & {65.16} \\
    FixMatch & \cmark & 78.01 & 68.93 & 87.79 & 73.75 & {77.12} & 77.30 & 68.67 & 80.49 & 73.32 & {74.94} \\
    FixMatch+RSC & \cmark & 79.57 & 71.32 & \textbf{90.97} & 67.77 & {77.41} & 76.43 & 67.06 & 87.34 & 63.80 & {73.66} \\
    \midrule
    \multicolumn{12}{c}{\emph{Semi-supervised domain generalization methods}} \\
    StyleMatch (\emph{ours}) & \cmark & \textbf{79.43} & \textbf{73.75} & {90.04} & \textbf{78.40} & \textbf{{80.41}} & \textbf{78.54} & \textbf{74.44} & \textbf{89.25} & \textbf{79.06} & \textbf{{80.32}} \\
    \bottomrule
    \end{tabular}
\end{table*}

\begin{table*}[t]
    \tabstyle{4pt}
    \caption{
    Domain generalization results in the low-data regime on OfficeHome. A: Art. C: Clipart. P: Product. R: Real world. $\bm{u}$: use unlabeled source data.
    }
    \label{tab:main_results_oh}
    \begin{tabular}{l | c | cccc >{\em}c | cccc >{\em}c}
    \toprule
    \multirow{2}{*}{Model} & \multirow{2}{*}{$\bm{u}$} & \multicolumn{5}{c|}{\# labels: 1950 (\emph{10 per class})} & \multicolumn{5}{c}{\# labels: 975 (\emph{5 per class})} \\
    & & A & C & P & R & {Avg} & A & C & P & R & {Avg} \\
    \midrule
    Full-Labels & - & 58.88 & 49.42 & 74.30 & 76.21 & {64.70} & 58.88 & 49.42 & 74.30 & 76.21 & {64.70} \\
    \midrule
    \multicolumn{12}{c}{\emph{Domain generalization methods}} \\
    Vanilla & \xmark & 50.11 & 43.50 & 65.11 & 69.65 & {57.09} & 45.76 & 39.97 & 60.04 & 63.77 & {52.38} \\
    CrossGrad & \xmark & 50.32 & 43.27 & 65.16 & 69.49 & {57.06} & 45.68 & 40.04 & 59.95 & 64.09 & {52.44} \\
    DDAIG & \xmark & 49.60 & 42.52 & 63.54 & 67.89 & {55.89} & 45.73 & 38.82 & 59.52 & 63.37 & {51.86} \\
    RSC & \xmark & 49.65 & 42.33 & 64.88 & 69.26 & {56.53} & 45.06 & 38.72 & 59.97 & 63.13 & {51.72} \\
    EISNet & \cmark & 51.16 & 43.33 & 64.72 & 68.36 & {56.89} & 47.32 & 40.07 & 59.33 & 62.59 & {52.33} \\
    \midrule
    \multicolumn{12}{c}{\emph{Semi-supervised learning methods}} \\
    MeanTeacher & \cmark & 49.92 & 43.42 & 64.61 & 68.79 & {56.69} & 44.65 & 39.15 & 59.18 & 62.98 & {51.49} \\
    EntMin & \cmark & 51.92 & 44.92 & \textbf{66.85} & \textbf{70.52} & {58.55} & 48.11 & 41.72 & \textbf{62.41} & 63.97 & {54.05} \\
    FixMatch & \cmark & 50.36 & 49.70 & 63.93 & 67.56 & {57.89} & 48.98 & 47.46 & 60.70 & 64.36 & {55.38} \\
    FixMatch+RSC & \cmark & 51.49 & 43.77 & 63.96 & 68.29 & {56.88} & 48.15 & 41.30 & 59.36 & 62.82 & {52.91} \\
    \midrule
    \multicolumn{12}{c}{\emph{Semi-supervised domain generalization methods}} \\
    StyleMatch (\emph{ours}) & \cmark & \textbf{52.82} & \textbf{51.60} & 65.31 & 68.61 & \textbf{{59.59}} & \textbf{51.53} & \textbf{50.00} & 60.88 & \textbf{64.47} & \textbf{{56.72}} \\
    \bottomrule
    \end{tabular}
\end{table*}

\section{Experiments}

\paragraph{SSDG Benchmarks}
We re-purpose two widely used DG datasets, PACS~\citep{li2017deeper} and OfficeHome~\citep{office_home}, for benchmarking SSDG methods. PACS consists of four distinct domains---art painting, cartoon, photo, and sketch---and contains 9,991 images of 7 classes in total. The domain shifts mainly concern image style changes. OfficeHome also has four domains: art, clipart, product, and real world. It contains more images than PACS, around 15,500 images of 65 classes, and more complex domain shifts, e.g., changes in image style, viewpoint, background, etc. See Fig.~\ref{fig:dataset_images} for example images from these two datasets.

\paragraph{Evaluation}
The common leave-one-domain-out protocol~\citep{li2017deeper} is adopted: three domains are used as the sources and the remaining one as the target. Note that only the source data are available for model training and the trained model is directly deployed in the target domain. Top-1 classification accuracy is reported. Two SSDG settings are designed. In the first setting, we randomly sample 10 images per class from each source domain as labeled data and treat the rest as unlabeled data. The second setting tests a more challenging scenario where only 5 labeled images are available for each class in each source domain. Results are averaged over five random splits.

\paragraph{Training Details}
Following the common practice~\citep{li2017deeper,zhou2020deep,huang2020self,zhou2021mixstyle}, the ImageNet-pretrained ResNet18~\citep{he2016deep} is used as the CNN backbone (for all models compared in this work). We randomly sample 16 images from each source domain to construct a minibatch, for labeled and unlabeled data respectively. Following FixMatch, the labeled minibatch is used for computing the labeled loss in Eq.~\eqref{eq:loss_labeled} while both labeled and unlabeled minibatches are used to compute the two unlabeled losses in Eq.~\eqref{eq:loss_strong} and \eqref{eq:loss_style}. The initial learning rate is set to 0.003 for the pretrained backbone and 0.01 for the randomly initialized stochastic classifier, both decayed by the cosine annealing rule. More implementation details can be found in our code.\footnote{\url{https://github.com/KaiyangZhou/ssdg-benchmark}.} All models are trained using a single Tesla V100 GPU. Our implementation is based on the public \texttt{Dassl.pytorch} toolbox~\citep{zhou2020domain}.\footnote{\url{https://github.com/KaiyangZhou/Dassl.pytorch}.}

\paragraph{Competitors}
We choose top-performing methods developed in two relevant fields, namely DG and SSL, for comparison. For DG, we choose CrossGrad~\citep{shankar2018generalizing}, DDAIG~\citep{zhou2020deep}, RSC~\citep{huang2020self}, EISNet~\citep{wang2020learning}, and also the vanilla model (also called ERM~\citep{gulrajani2021in} or Deep-All~\citep{li2017deeper} in the literature). These DG methods can only use labeled data for training except EISNet, whose design based on self-supervised learning allows the model to use unlabeled data. For SSL, we choose MeanTeacher~\citep{tarvainen2017mean}, Entropy-Minimization (EntMin)~\citep{grandvalet2004semi}, and FixMatch~\citep{sohn2020fixmatch}, which have been widely used as baselines in the SSL literature.

\begin{figure*}
	\centering
	\includegraphics[width=.95\textwidth]{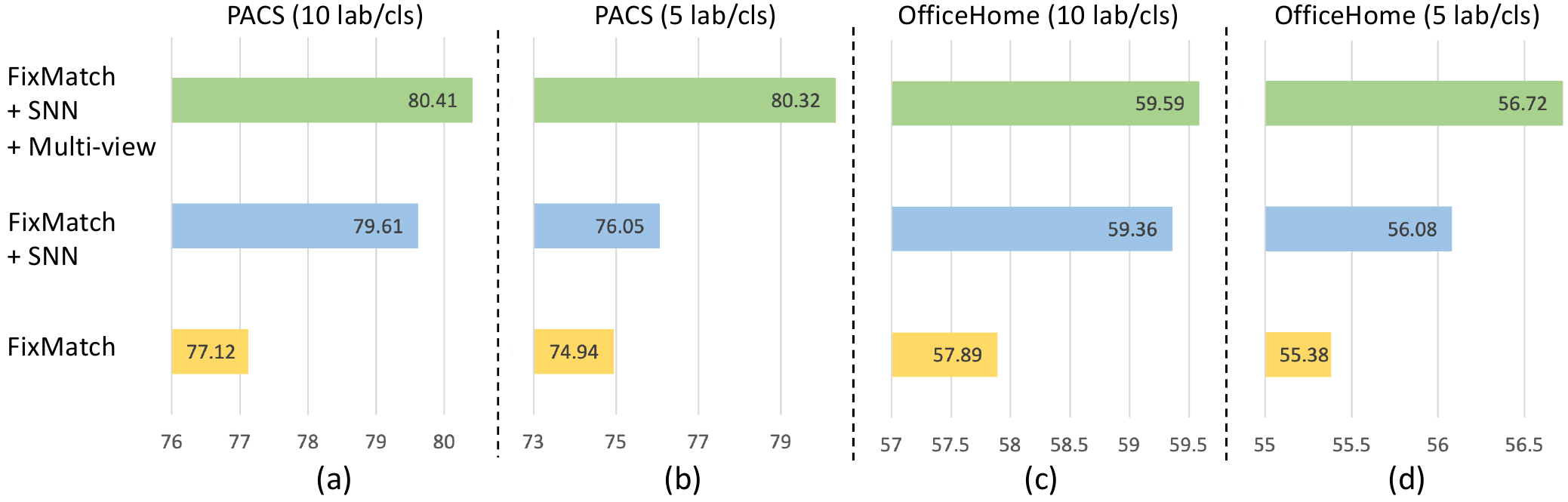}
	\caption{Ablation study on the two core components in StyleMatch: the SNN classifier (Fig.~\ref{fig:method}(a)) and the multi-view consistency learning (Fig.~\ref{fig:method}(b)). Clearly, the two components are complementary to each other and together bring significant improvements over the baseline model (i.e., FixMatch).}
	\label{fig:main_ablation}
\end{figure*}

\subsection{Main Results}
The results on PACS and OfficeHome are presented in Table~\ref{tab:main_results_pacs} and \ref{tab:main_results_oh}, respectively. Full-Labels refers to the vanilla model trained using all labels in the source data, which can be seen as the pseudo oracle.

From both tables we can observe that the DG methods that rely purely on the limited labeled data do not work well and hardly beat the vanilla model. This is not unexpected as most DG algorithms are subject to the assumption that rich, fully-labeled multi-source data are available for model training. Among the DG methods, EISNet's performance stands out since it takes advantage of the unlabeled data.

The SSL methods generally outperform the DG methods on both datasets. It is worth noting that OfficeHome is more challenging than PACS as it contains more classes (65 vs 7) and cluttered images. Therefore, a small improvement on OfficeHome can be considered significant. FixMatch clearly outperforms the two peers in the same group, i.e., MeanTeacher and EntMin.

We also combine FixMatch with RSC---the latter is based on suppressing dominant features with large gradients so the model is forced to use more features for prediction, and currently has the best reported results in the standard DG setting~\citep{huang2020self}. Clearly, FixMatch+RSC does not bring significant improvements, suggesting that \emph{a naive combination of the best techniques from both worlds is insufficient for solving SSDG}.

Our StyleMatch shows clear advantages over all competitors including FixMatch, which justify the effectiveness of our designs for SSDG. On PACS, when the target domains are art painting (A) and sketch (S), StyleMatch even surpasses Full-Labels with a clear margin. On OfficeHome, there is still room for improvement when comparing StyleMatch with the ``oracle'' Full-Labels. It is noteworthy that when moved from the 10-labels to 5-labels setting, StyleMatch's average gain over the vanilla model is increased significantly: from around 17\% to 25\% on PACS, and around 2\% to 4\% on OfficeHome. The results suggest that StyleMatch is highly competent in extreme low-data scenarios.

\subsection{Ablation Study and Analysis}

\begin{figure*}
	\centering
	\includegraphics[width=.95\textwidth]{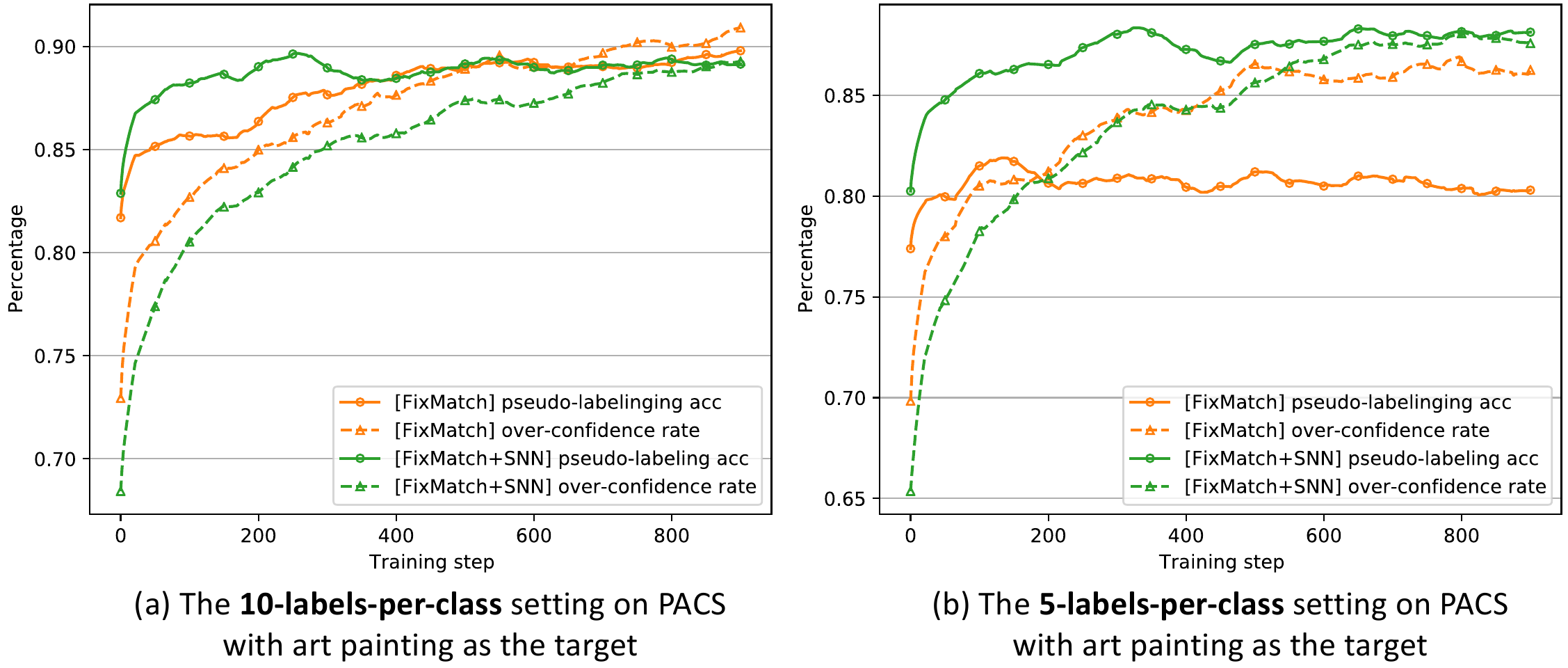}
	\caption{Pseudo-labeling accuracy (\emph{solid+circle}) vs overconfidence rate (\emph{dashed+triangle}). Without the SNN classifier, the model suffers from severe overfitting, which is reflected in the overconfidence rate overshooting the pseudo-labeling accuracy.}
	\label{fig:snn_effect_pseudolab}
\end{figure*}

\paragraph{Key Components}
We conduct a comprehensive ablation study on the two key components in StyleMatch, i.e., the SNN classifier (Fig.~\ref{fig:method}(a)) and the multi-view consistency learning framework (Fig.~\ref{fig:method}(b)). We repeat the experiments on PACS and OfficeHome by sequentially adding these two components to FixMatch---the base model on top of which StyleMatch is built. Fig.~\ref{fig:main_ablation} shows the results of this study with a focus on the average accuracy over all target domains. It is clear that the SNN classifier contributes an around 2\%/1\% increase to the performance on PACS/OfficeHome, and the multi-view design further boosts the performance. The improvements brought by the multi-view design are higher in the lower-data setting on both datasets, suggesting that this design is essential when dealing with extremely scarce labels.

\paragraph{Stochastic Classifier Reduces Overfitting}
To understand how the stochastic classifier improves learning, we compare FixMatch+SNN with FixMatch using two metrics: pseudo-labeling accuracy and overconfidence rate, which are measured for each minibatch data received at each training step. The first metric measures the accuracy of pseudo-labels, while the overconfidence rate counts how many pseudo-labels in a minibatch pass the confidence threshold. Ideally, we do not want the overconfidence rate to climb above the pseudo-labeling accuracy as this would mean the network predicts excessive incorrect pseudo-labels with high confidence, which hurt generalization~\citep{zhang2017understanding}. Fig.~\ref{fig:snn_effect_pseudolab} shows the comparisons. In (a), the overconfidence rate of FixMatch+SNN steadily increases and eventually converges to a similar level as the pseudo-labeling accuracy. In contrast, without SNN, the overconfidence rate overshoots the pseudo-labeling accuracy in the middle of training. In (b), the overfitting issue for FixMatch intensifies---the overconfidence rate outpaces the pseudo-labeling accuracy at the early training stage and the pseudo-labeling accuracy stops improving since then. In contrast, the curves for FixMatch+SNN look much healthier.

\paragraph{Augmentation Methods}
To have an in-depth understanding of the roles of the augmentation methods, we evaluate three variants of StyleMatch: $T_{strong}$-only, $T_{style}$-only, and $T_{strong}\!+\!T_{style}$. $T_{strong}$-only and $T_{style}$-only are based on the two-view consistency learning paradigm and $T_{strong}\!+\!T_{style}$ refers to the final model. Table~\ref{tab:Tstrong_Tstyle} shows the results of this ablation study on PACS. We observe that 1) $T_{strong}$ is more suitable than $T_{style}$ to be used in the two-view consistency learning framework, and 2) combining these two augmentation methods leads to a much better performance, which justifies their complementarity.

\paragraph{Number of Source Domains}
So far the experiments are based on the commonly used three-source setting. How does the approach fare when there are less sources? To answer this question, we reduce the number of source domains from three to two and one, and conduct the experiments on PACS. When using one domain as the target, the results are the average over all possible scenarios with different combinations of sources, each still following the five random splits protocol. Table~\ref{tab:n_source} details the results where StyleMatch is compared with FixMatch. Note that when $K\!=\!1$ (single-source case), StyleMatch mixes the image style between random instances from the same domain. The results show that StyleMatch outperforms FixMatch with a clear margin in all scenarios, even in the single-source case---this means that mixing instance-level style also helps, which is also observed in a recent work that mixes instance-level feature statistics~\citep{zhou2021mixstyle}. By increasing $K$ from 2 to 3, StyleMatch gains 5.91\% (from 74.50\% to 80.41\%) and 8.37\% (from 71.95\% to 80.32\%) respectively in the 10- and 5-labels settings, while FixMatch's gains are 5.7\% (from 71.42\% to 77.12\%) and 6.42\% (from 68.52\% to 74.94\%) respectively, which are smaller. This suggests that StyleMatch can better handle heterogeneous data.

\begin{table}[t]
    \tabstyle{5pt}
    \caption{An analysis of augmentation methods. The proposed design, which combines $T_{strong}$ with $T_{style}$ proves to be the optimal choice.}
    \label{tab:Tstrong_Tstyle}
    \begin{tabular}{l c c}
    \toprule
    & \multicolumn{2}{c}{PACS} \\
    StyleMatch's variants & 10 lab/cls & 5 lab/cls \\
    \midrule
    $T_{strong}$-only & 79.61 & 76.05 \\
    $T_{style}$-only & 72.61 & 69.72 \\
    $T_{strong}\!+\!T_{style}$ & \textbf{80.41} & \textbf{80.32} \\
    \bottomrule
    \end{tabular}
\end{table}

\begin{table}[t]
    \tabstyle{3pt}
    \caption{Impact on number of sources ($K$). StyleMatch consistently improves upon FixMatch in all scenarios.}
    \label{tab:n_source}
    \begin{tabular}{l ccc ccc}
    \toprule
    & \multicolumn{6}{c}{PACS} \\
    & \multicolumn{3}{c}{10 lab/cls} & \multicolumn{3}{c}{5 lab/cls} \\ \cmidrule(lr){2-4} \cmidrule(lr){5-7}
    & $K\!=\!1$ & $K\!=\!2$ & $K\!=\!3$ & $K\!=\!1$ & $K\!=\!2$ & $K\!=\!3$ \\ \midrule
    FixMatch & 53.55 & 71.42 & 77.12 & 49.91 & 68.52 & 74.94 \\
    StyleMatch & \textbf{57.29} & \textbf{74.50} & \textbf{80.41} & \textbf{52.24} & \textbf{71.95} & \textbf{80.32} \\
    \bottomrule
    \end{tabular}
\end{table}

\section{Conclusion}
Model generalization and data-efficiency are two critical problems for real-world vision applications but have been studied separately in the literature. This work explores a unification of the two problems, called semi-supervised domain generalization (SSDG), which poses unique challenges and requires specific designs. Through extensive experiments in the two newly proposed benchmarks, we show that state-of-the-art DG or SSL methods could not suffice to solve the problem, nor a naive combination of them. The proposed approach, despite having concise designs, demonstrates significant gains in the experiments and can serve as a strong baseline for future research.


%
%

\bibliographystyle{spbasic}      
\bibliography{ref}   

\end{document}